\documentclass[10.7pt,]{article}

\usepackage[letterpaper, margin=2.54cm, top=2.54cm]{geometry}
\usepackage{algorithm}
\usepackage{algorithmic}
\usepackage{subcaption}
\usepackage[super,comma,sort&compress]{natbib}
\usepackage{lmodern}
\usepackage{authblk} 
\usepackage{amssymb,amsmath}
\usepackage{wrapfig}
\usepackage{graphicx,grffile}
\usepackage[labelfont=bf,labelsep=period]{caption}
\usepackage{ifxetex,ifluatex}
\usepackage{fixltx2e} 
\usepackage[T1]{fontenc}           
\usepackage[utf8]{inputenc}        
\usepackage{newtxtext,newtxmath}   
\ifnum 0\ifxetex 1\fi\ifluatex 1\fi=0 
  \usepackage[T1]{fontenc}
  \usepackage[utf8]{inputenc}

\IfFileExists{upquote.sty}{\usepackage{upquote}}{}
\IfFileExists{microtype.sty}{%
	\usepackage{microtype}
	\UseMicrotypeSet[protrusion]{basicmath} 
}{}

\usepackage{sectsty}
\allsectionsfont{\fontsize{10}{10}\selectfont}
\subsectionfont{\itshape\bfseries\fontsize{10}{10}\selectfont}
\subsubsectionfont{\normalfont\itshape}

\makeatletter
\renewcommand\subsubsection{\@startsection{subsubsection}{3}{\z@}%
	{-3.25ex\@plus -1ex \@minus -.2ex}%
    {-1.5ex \@plus -.2ex}
    {\normalfont\itshape}}
\makeatother

\makeatletter 
\renewcommand\@biblabel[1]{#1.} 
\makeatother %

\usepackage{etoolbox}
\makeatletter
\patchcmd{\@maketitle}{\LARGE}{\bfseries\fontsize{15}{16}\selectfont}{}{}
\makeatother

\makeatletter
\def\maxwidth{\ifdim\Gin@nat@width>\linewidth\linewidth\else\Gin@nat@width\fi}
\def\maxheight{\ifdim\Gin@nat@height>\textheight\textheight\else\Gin@nat@height\fi}
\makeatother

\setkeys{Gin}{width=\maxwidth,height=\maxheight,keepaspectratio}
\ifx\paragraph\undefined\else
\let\oldparagraph\paragraph
\renewcommand{\paragraph}[1]{\oldparagraph{#1}\mbox{}}
\fi
\ifx\subparagraph\undefined\else
\let\oldsubparagraph\subparagraph
\renewcommand{\subparagraph}[1]{\oldsubparagraph{#1}\mbox{}}
\fi
\usepackage{hyperref}
\hypersetup{
	unicode=true,
	pdftitle={Automating Adjudication of Cardiovascular Events Using Large Language Models},
	pdfauthor={Sonish Sivarajkumar, Min Jiang},
	pdfkeywords={clinical trials, cardiovascular events, adjudication, large language models, artificial intelligence},
	pdfborder={0 0 0},
	breaklinks=true
}
\title{\vspace{-2em} Automating Adjudication of Cardiovascular Events Using Large Language Models}

\author[ ]{\bf\fontsize{13}{14}\selectfont 
Sonish Sivarajkumar, BS\textsuperscript{1,2}, 
Kimia Ameri, PhD\textsuperscript{1}, 
Chuqin Li, PhD\textsuperscript{1}, 
Yanshan Wang, PhD\textsuperscript{2}, 
Min Jiang, PhD\textsuperscript{1}\thanks{Corresponding author: Min Jiang, PhD, \texttt{jiang\_min@lilly.com}} \vspace{-.7em}
}

\affil[1]{\bf\fontsize{13}{14}\selectfont Advanced Analytics and Data Sciences, Eli Lilly and Company, USA}
\affil[2]{\bf\fontsize{13}{14}\selectfont Intelligent Systems Program, School of Computing and Information, University of Pittsburgh, USA}

\date{} 

\begin{document}
\maketitle
\vspace{-4em} 

\section{Abstract}\label{abstract}
\emph{Cardiovascular events, such as heart attacks and strokes,  remain a leading cause of mortality globally, necessitating meticulous monitoring and adjudication in clinical trials. This process, traditionally performed manually by clinical experts, is time-consuming, resource-intensive, and prone to inter-reviewer variability, potentially introducing bias and hindering trial progress. This study addresses these critical limitations by presenting a novel framework for automating the adjudication of cardiovascular events in clinical trials using Large Language Models (LLMs). We developed a two-stage approach: first, employing an LLM-based pipeline for event information extraction from unstructured clinical data and second, using an LLM-based adjudication process guided by a Tree of Thoughts approach and clinical endpoint committee (CEC) guidelines. Using cardiovascular event-specific clinical trial data, the framework achieved an F1-score of 0.82 for event extraction and an accuracy of 0.68 for adjudication. Furthermore, we introduce the CLEART score, a novel, automated metric specifically designed for evaluating the quality of AI-generated clinical reasoning in adjudicating cardiovascular events. This approach demonstrates significant potential for substantially reducing adjudication time and costs while maintaining high-quality, consistent, and auditable outcomes in clinical trials. The reduced variability and enhanced standardization also allows for faster identification and mitigation of risks associated with cardiovascular therapies.}

\section{Introduction}\label{introduction}

Cardiovascular (CV) events—such as heart attacks, strokes, and other conditions that damage the heart muscle or disrupt blood flow to the brain—remain a leading cause of morbidity and mortality worldwide, with millions of new cases each year\cite{mensah2019global}. These events serve as pivotal markers in assessing the efficacy and safety of novel treatments during clinical trials; therefore, accurate identification and adjudication of CV events are essential for ensuring valid trial outcomes\cite{hicks20182017}. However, adjudicating these events is particularly time-consuming due to the reliance on unstructured clinical documents (e.g., physician notes, discharge summaries, imaging reports) that must be carefully reviewed by multiple experts for confirmation. Human-driven chart review faces substantial challenges, including inconsistent documentation practices, missing data, complex clinical terminology, and the risk of inter-reviewer variability—a key source of bias that can potentially compromise the reliability of trial results\cite{hill2005bias,khan2023endpoint,pogue2009evaluating}. These factors can introduce bias, slow down the drug development process, and increase the overall cost of clinical trials\cite{movsas2003quality}. 

Despite these complexities, there has been steady progress toward automating aspects of CV event detection using Natural Language Processing (NLP) methods. Several studies have demonstrated the potential of NLP for identifying specific cardiovascular diagnoses or complications from clinical notes—for example, detecting heart failure or related diagnoses through rule-based systems, deep learning models, or machine learning pipelines\cite{latif2020implementation,sevakula2020state}. However, these approaches typically focus on a narrow set of CV events (e.g., myocardial infarctions or heart failure) and do not fully automate the more complex, multi-step process of event adjudication according to clinical endpoint committee (CEC) guidelines.

To address these gaps, we propose leveraging recent advances in Large Language Models (LLMs) to handle the substantial complexity inherent in adjudicating CV events from large volumes of clinical text. LLMs have shown remarkable promise in various healthcare applications, including natural language understanding and information extraction from complex clinical texts\cite{sivarajkumar2023healthprompt, sivarajkumar2024empirical}. By encoding CEC guidelines directly into an LLM-driven pipeline, we aim to standardize the decision-making procedure in a way that improves both consistency and transparency, offering an auditable record of each AI-generated decision. 

In this study, we introduce a novel framework for automating the adjudication of cardiovascular deaths, featuring a two-stage process: an LLM-based event information extraction phase, followed by a Tree of Thoughts approach to synthesize and classify events as CV or non-CV in alignment with established guidelines. Our evaluation underscores how such a framework can significantly reduce the manual burden on clinical endpoint committees, accelerate trial operations, and maintain a high degree of consistency in adjudication decisions.

\section{Methods}

\subsection{Design}
We introduce a novel two-stage framework for automating the adjudication of CV deaths in clinical trials. This approach was developed to address the significant challenges associated with processing large volumes of unstructured clinical data while ensuring strict adherence to established CEC guidelines\cite{seltzer2015centralized}. The framework consists of two primary stages: Event Information Extraction and LLM-based Adjudication.

The first stage, Event Information Extraction, serves as the foundation of our framework. It is designed to efficiently process and structure the vast amounts of unstructured clinical data typically encountered in large-scale clinical trials, such as electronic health records (EHRs), medical notes, and laboratory results. This stage employs advanced natural language processing (NLP) techniques, specifically an LLM-based few-shot information extraction approach\cite{Beltagy2019}. This approach is used to identify and extract relevant clinical events from electronic health records (EHRs), including CV diagnoses, procedures, medications, and dates. The extracted information forms a structured dataset that serves as input for the subsequent adjudication stage.

The second stage, LLM-based Adjudication, represents the core of our automated decision-making process. This stage utilizes a sophisticated Tree of Thoughts (ToT) approach, which enables the framework to explore multiple reasoning paths simultaneously, mimicking the complex decision-making process of human experts\cite{yao2024tree}. By incorporating CEC guidelines directly into the LLM’s reasoning process, we ensure that the automated adjudication closely aligns with established clinical standards. The framework is designed to be flexible, allowing for the use of any LLM according to specific requirements, including both API-based models like GPT-4 and open-source models like LLaMA-3\cite{dubey2024llama}. The modular design makes the system adaptable to different institutional and trial-specific requirements, offering a scalable solution for automated adjudication.

\begin{figure}[h]
\centering
\includegraphics[width=0.9\textwidth]{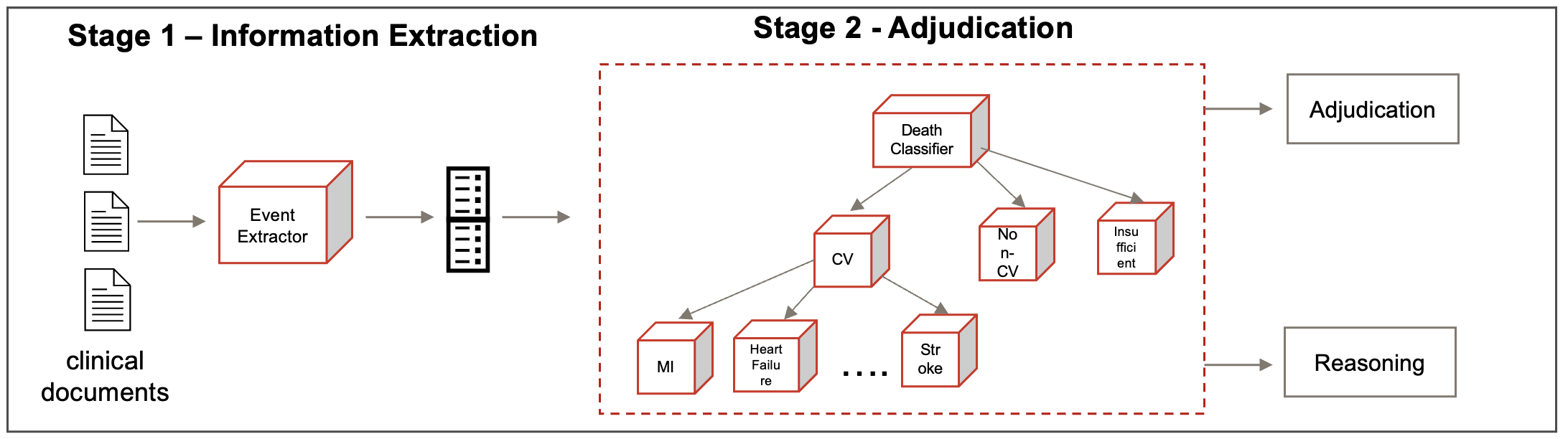}
\caption{Agent-based adjudication framework}
\label{fig:overall}
\end{figure}

The study utilized clinical trial data from Eli Lilly and Company, focusing specifically on cardiovascular death events. This setting provided a rich source of real-world clinical data, encompassing the complexity and variability typical in large-scale clinical trials.

\subsection{Event Information Extraction}

We developed an LLM-based pipeline to extract relevant clinical events from unstructured electronic health records (EHRs). This stage is crucial for reducing the vast amount of clinical data into a structured format that can be efficiently processed by the adjudication model. Our approach leverages a few-shot learning paradigm powered by an LLM, a methodology necessitated by the absence of ground truth data for training—a common challenge in specialized clinical domains\cite{oniani2023few}. This prompt-based few-shot learning technique has proven effective in numerous clinical NLP studies, particularly in scenarios with low or zero training data availability\cite{sivarajkumar2024empirical}.

Our extraction pipeline, illustrated in Figure 2, consists of four key steps: Sentence Segmentation, Tokenization, Entity Detection, and Relation Detection. The process begins with Sentence Segmentation, which divides the raw text of each clinical document into individual sentences. This is followed by Tokenization, breaking down each sentence into its constituent words or tokens. Both these steps were performed using the NLTK library. Next, an LLM identifies relevant clinical events within the tokenized text, including CV event names, negation, and temporal information.

\begin{figure}[h]
\centering
\includegraphics[width=1\textwidth]{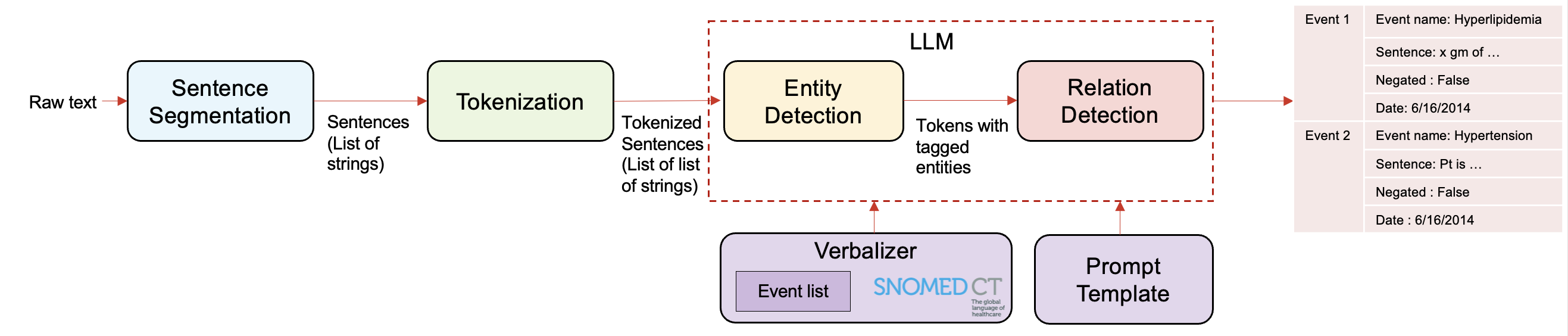}
\caption{Event extraction pipeline: LLM-based zero-shot clinical information extraction}
\label{fig:event_extraction}
\end{figure}

The LLM-based approach replaces traditional rule-based or machine learning models with a more flexible and powerful language model. The process can be described by the following equation:
\begin{equation}
E = f_{LLM}(D, P, V)
\end{equation}
Where $E$ is the extracted event information, $f_{LLM}$ is the LLM function, $D$ is the input clinical document,  $P$ is the prompt template guiding the extraction, and $V$ is the verbalizer(label list).  

We enhanced our verbalizer (V) with synonyms from SNOMED CT to improve entity recognition and categorization\cite{gaudet2021use}.These terms were derived from the standardized definitions of cardiovascular deaths as outlined by established CEC guidelines, ensuring that all relevant clinical terms are accurately captured. For "Myocardial Infarction," the verbalizer includes terms such as Heart Attack, Cardiac Infarction, AMI, STEMI, and NSTEMI. This augmentation enables accurate identification of clinical events across various terminologies. 

\begin{figure}[h]
\centering
\includegraphics[width=1\textwidth]{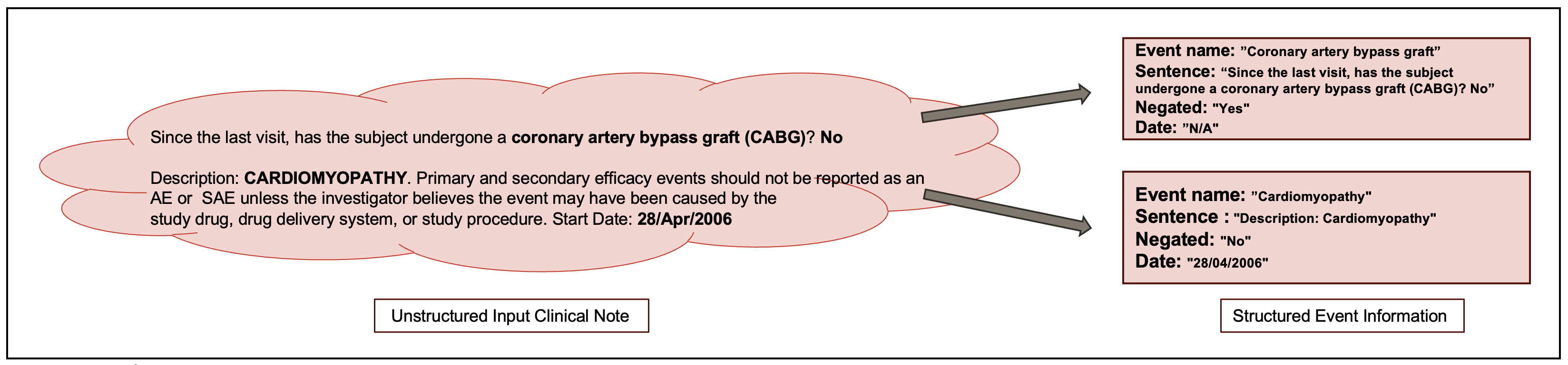}
\caption{Event information extraction - An example}
\label{fig:example_event_extraction}
\end{figure}

For each event, our system extracts four key elements, as shown in figure \ref{fig:example_event_extraction}:

\begin{itemize}
\item the event name (a standardized term for the clinical event),
\item associated sentence(s) (the specific textual context in which the event was mentioned), 
\item negation status (whether the event is positively affirmed or negated in the text), and 
\item the date of the event (temporal information associated with the event, when available).
\end{itemize}

\subsection{LLM-based Adjudication}
We implemented a Tree of Thoughts (ToT) approach using GPT-4 to classify deaths as cardiovascular or non-cardiovascular. GPT-4 was chosen for its advanced reasoning capabilities and capacity to handle complex, multi-step prompts, which would more closely mimic the nuanced decision-making process employed by human adjudicators. This process incorporated CEC guidelines into the LLM's decision-making process. The ToT approach can be represented as:

\begin{equation}
(A, R) = ToT_{LLM}(E, G)
\end{equation}

Where $A$ is the adjudication decision, $R$ is the reasoning, $ToT_{LLM}$ is the Tree of Thoughts LLM function, $E$ is the extracted event information, and $G$ is the CEC guideline.

\begin{figure}[h]
    \centering
    \begin{subfigure}[b]{0.8\textwidth}
        \centering
        \includegraphics[width=\textwidth]{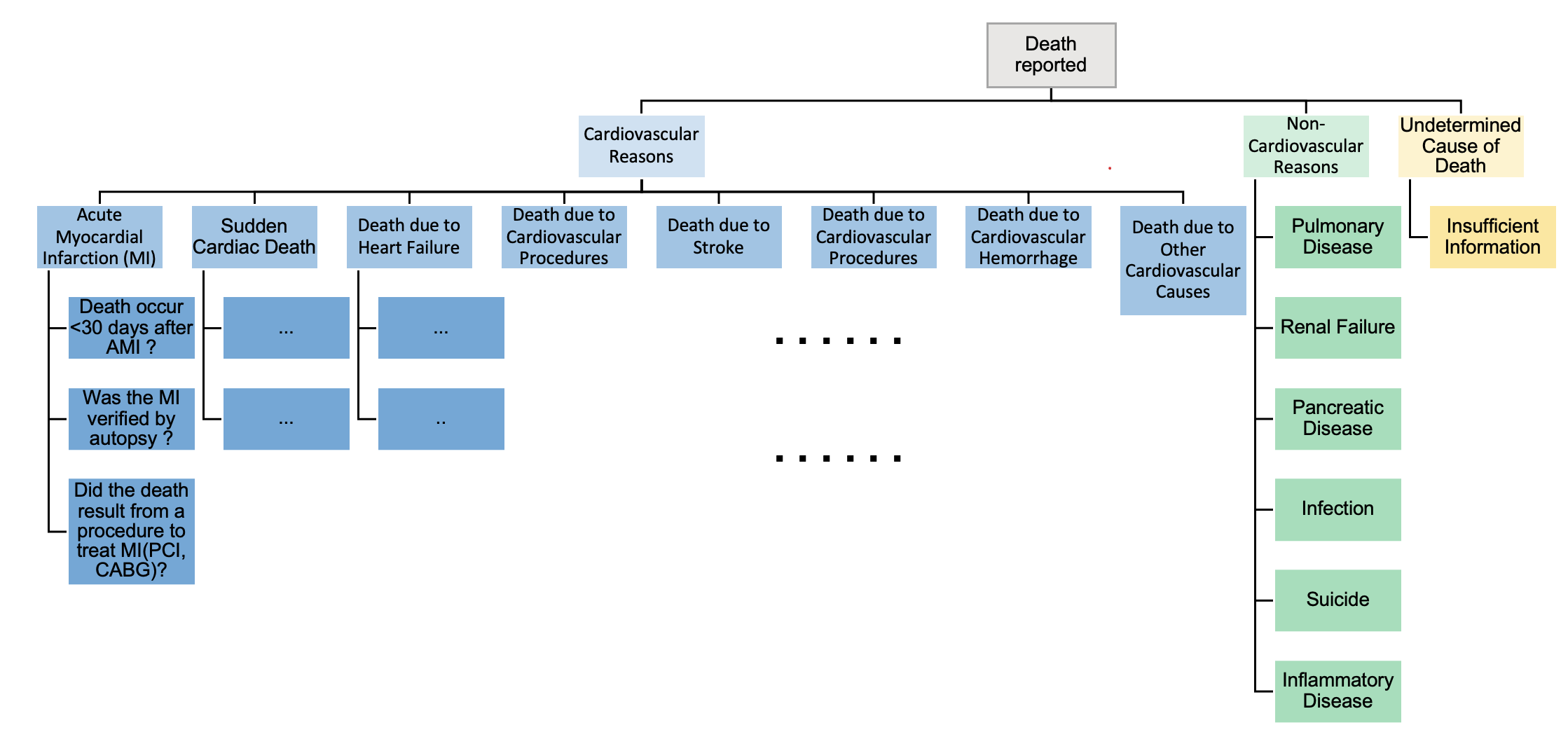}
        \caption{}
        \label{fig:adjudicationprocess_a}
    \end{subfigure}
    \vfill
    \begin{subfigure}[b]{0.5\textwidth}
        \centering
        \includegraphics[width=\textwidth]{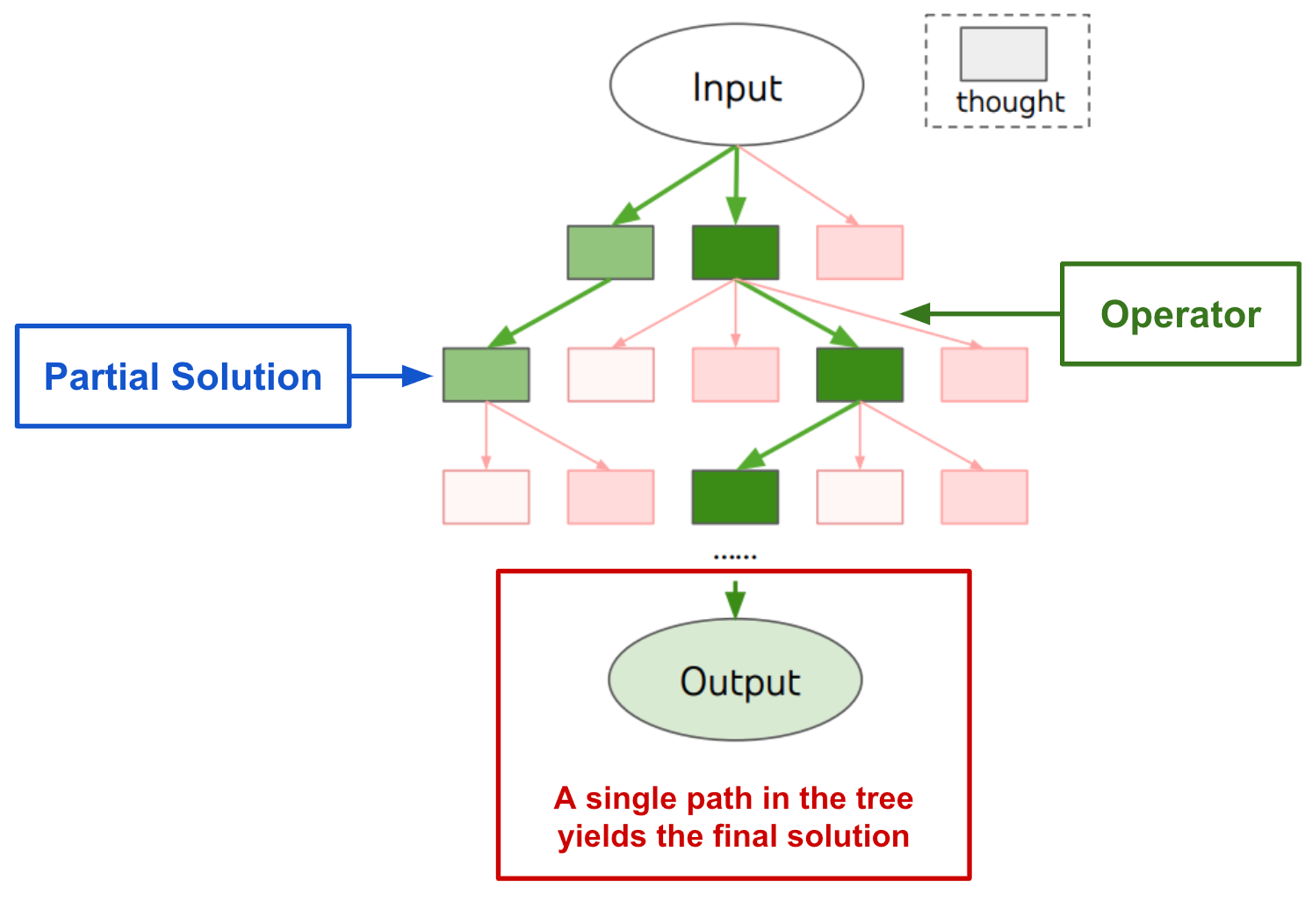}
        \caption{}
        \label{fig:adjudicationprocess_b}
    \end{subfigure}
    \caption{(a) LLM-based adjudication process; (b) Tree of thought process}
    \label{fig:adjudicationprocess_combined}
\end{figure}

The algorithm can be described as follows:

\begin{algorithm}
\caption{Tree of Thoughts for Cardiovascular Death Adjudication}
\begin{algorithmic}[1]
\REQUIRE Input E (extracted data), LLM $p_\theta$, guidelines G
\STATE $is\_dead \gets \text{LLM}(p_\theta, E)$ \COMMENT{Step 1: Determine if patient is deceased}
\IF{$is\_dead$}
    \STATE $acute\_mi \gets \text{LLM}(p_\theta, E)$
    \STATE $sudden\_cardiac\_death \gets \text{LLM}(p_\theta, E)$
    \STATE $heart\_failure \gets \text{LLM}(p_\theta, E)$
    \STATE $stroke \gets \text{LLM}(p_\theta, E)$
    \STATE $cv\_procedure \gets \text{LLM}(p_\theta, E)$
    \STATE $cv\_hemorrhage \gets \text{LLM}(p_\theta, E)$
    \STATE $other\_cv\_causes \gets \text{LLM}(p_\theta, E)$
    \STATE $non\_cv\_causes \gets \text{LLM}(p_\theta, E)$
    \STATE $undetermined \gets \text{LLM}(p_\theta, E)$
    \STATE $final\_reasoning \gets \text{LLM}(p_\theta, E, all\_previous\_states)$
\ENDIF
\RETURN $final\_reasoning$
\end{algorithmic}
\end{algorithm}

This algorithm operates through a series of carefully designed prompts and chains, each representing a node in the thought tree. The process begins with an initial step to determine if the patient is deceased, followed by a series of specific prompts to classify the cause of death according to the CEC guidelines. 

The thought tree for our cardiovascular death adjudication process consists of multiple levels, each representing a different aspect of the decision-making process. The root node contains the extracted event information. The first level determines whether the patient is deceased. Subsequent levels explore different potential causes of death, including Acute Myocardial Infarction, Sudden Cardiac Death, Heart Failure, Stroke, Cardiovascular Procedures, Cardiovascular Hemorrhage, Other Cardiovascular Causes, Non-Cardiovascular Causes, and Undetermined causes. Each node in the tree represents a specific prompt designed to evaluate a particular aspect of the case. For instance, the prompt for Acute Myocardial Infarction considers factors such as the timing of death relative to the MI, verification by diagnostic criteria or autopsy, and whether the death resulted from a procedure to treat MI. New nodes can be added to the tree to represent additional decision points or criteria, allowing the system to evolve with changing medical knowledge and adjudication practices.

The final step in the ToT adjudication process involves a consolidation prompt that synthesizes the information from all explored paths. This prompt considers the outcomes of each potential cause of death and generates a final reasoning that explains the adjudication decision. This final reasoning provides a clear, traceable path from the input data to the final classification, enhancing the interpretability of the AI-generated decision. By structuring our adjudication process as a Tree of Thoughts, we enable the LLM to consider multiple aspects of the case in a systematic manner, similar to how a human expert would approach the problem. This approach allows for nuanced decision-making that can handle the complexities and ambiguities often present in cardiovascular death cases. 


\section{Evaluation}

Our evaluation strategy for the automated cardiovascular death adjudication framework was designed to comprehensively assess both the accuracy of our system and the quality of its reasoning. This multi-faceted approach addresses the complex nature of the task and evaluates different aspects of the system's performance.

\subsection{Performance Metrics}

For the Event Information Extraction stage, we employed standard information retrieval metrics to assess the effectiveness of our few-shot learning approach using an LLM-based pipeline. We calculated precision to determine the proportion of extracted events that were correctly identified, providing insight into the system's accuracy. Recall was measured to assess the proportion of all relevant events in the text that were successfully extracted, indicating the system's completeness. To balance these two aspects, we computed the F1-score, which provides a single, harmonized measure of the system's performance.

In addition to these overall performance metrics, we evaluated two attributes of event extraction: negation detection and date extraction. We assessed the accuracy of the system in determining whether events were affirmed or negated, and in associating them with the correct timeframes.

For the LLM-based Adjudication stage, our primary focus was on the accuracy of the adjudication decisions compared to expert adjudication. This involved a direct comparison between the system's classification of deaths (as either cardiovascular or non-cardiovascular) and the gold standard established by human expert decisions. For the binary classification of cardiovascular deaths vs. non-cardiovascular deaths, we report accuracy, defined as the total number of correct adjudications (true positives plus true negatives) divided by the total number of adjudications. We emphasize that accuracy is distinct from precision because it incorporates both correctly identified positive and negative outcomes, making it appropriate for final classification results.

We implemented our Tree of Thoughts approach using GPT-4 and compared its performance against the baseline method: a simpler Summarizer + Adjudicator approach. A single-pass LLM approach baseline was not feasible as some of the clinical notes exceeded the context window of most LLMs.

\subsection{CLEART Score: Automated Quantitative Confidence Score}

For evaluating the quality of adjudication reasoning, especially in the absence of ground truth, we developed a novel metric, the CLEART score. This novel metric provides a comprehensive assessment of the reasoning process, focusing on six key aspects that are crucial in clinical decision making\cite{tam2024framework}, as shown in Table \ref{tab:cleart_criteria}.

\begin{table}[h]
\centering
\begin{tabular}{|l|l|}
\hline
\textbf{Criterion} & \textbf{Description} \\
\hline
Clarity & Clarity of reasoning without ambiguities \\
Logical consistency & Logical consistency without contradictions \\
Evaluation details & Inclusion of specific clinical reasoning and key details \\
Adherence to guidelines & Strict adherence to provided guidelines \\
Relevance & Correct use of diagnostic criteria/autopsy findings \\
Timeline accuracy & Correct identification of relevant time frames \\
\hline
\end{tabular}
\caption{CLEART Score Criteria}
\label{tab:cleart_criteria}
\end{table}

Each of these criteria is scored on a binary scale, where 0 indicates that the criterion was not met, and 1 indicates that it was satisfactorily met. The final CLEART score is calculated as the average of these six individual scores, providing a single, comprehensive measure of the quality of the framework's reasoning:

\begin{equation}
CLEART = \frac{1}{6} \sum_{i=1}^{6} C_i
\end{equation}

where $C_i$ represents the score (0 or 1) for each of the six criteria.

To generate the CLEART score, we employed another LLM as an automated evaluator. This LLM is provided with the AI-generated reasoning and a rubric detailing the CLEART criteria. It then assesses the reasoning against each criterion, providing a binary score and a brief justification for each. This approach allows us to generate quantitative confidence scores for the framework's reasoning without relying on human experts, addressing the lack of ground truth for reasoning evaluation.

The CLEART score serves multiple important functions in our evaluation framework. It provides a quantitative measure of the quality of the generated reasoning, allowing for objective comparison between different models or approaches. The score helps identify specific areas of strength or weakness in the reasoning process, guiding our efforts in refining and improving the adjudication system. It also offers a level of interpretability and transparency to the framework's decision-making process, which is crucial for building trust in clinical applications where the stakes are high and the reasoning behind decisions is as important as the decisions themselves.


\section{Results}

\subsection{Event Information Extraction}
The CV event extraction stage demonstrated strong performance in identifying and structuring relevant clinical events from unstructured clinical notes.
The system achieved a precision of 0.96, indicating high accuracy in the events it identified. While the recall was lower at 0.71, it still suggests that the majority of relevant events were successfully captured. Overall f1 score 0.82 shows that the few-shot learning approach is effective in event information extraction, which is in synergy with some of the existing studies \cite{sivarajkumar2024extraction, sivarajkumar2024empirical}. Given its foundational role in providing the evidence needed for adjudication, high-quality information extraction, particularly in accurately capturing both entities and associated attributes, is paramount; there is still room to improve its effectiveness by further exploring attribute extraction.

Attribute-level accuracy was also high, with negation detection achieving 0.86 accuracy and date extraction achieving 0.81 accuracy. On average, the system extracted 49 events per patient, demonstrating its ability to comprehensively capture relevant clinical information.

\subsection{LLM-based Adjudication}

We evaluated different approaches for the adjudication task to determine the most effective method for classifying cardiovascular deaths. Table \ref{tab:adjudication_accuracy} presents the accuracy of each model:

\begin{table}[h]
\centering
\begin{tabular}{|l|c|}
\hline
\textbf{Model} & \textbf{Accuracy} \\
\hline
Tree of Thought – GPT 4 & 0.68 \\
Summarizer + Adjudicator (GPT 4) & 0.60 \\
Tree of Thought – LLAMA 3 & 0.65 \\
\hline
\end{tabular}
\caption{Adjudication Accuracy by Model}
\label{tab:adjudication_accuracy}
\end{table}

The Tree of Thought approach using GPT-4 demonstrated the highest accuracy at 0.68. This superior performance suggests that the more complex reasoning process enabled by the Tree of Thought methodology is beneficial for this task. Interestingly, the Tree of Thought approach implemented with open-source LLAMA 3 model achieved an accuracy of 0.65, notably close to the performance of GPT-4. The Summarizer + Adjudicator approach achieved a lower accuracy of 0.60. While summarization can help in managing large amounts of information, it may lose crucial details necessary for accurate adjudication.

\subsection{CLEART Score}
Table \ref{tab:cleart_scores} shows the average score of each criterion evaluated for 100 patients. Average scores indicate which criterion is more helpful for evaluating the rationale of the decision.

\begin{table}[h]
\centering
\begin{tabular}{|l|c|}
\hline
\textbf{Criterion} & \textbf{Average Score} \\
\hline
Clarity & 0.69 \\
Logical consistency & 0.98 \\
Evaluation details & 0.71 \\
Adherence to guidelines & 0.96 \\
Relevance & 0.55 \\
Timeline accuracy & 0.31 \\
\hline
Overall CLEART score & 0.67 \\
\hline
\end{tabular}
\caption{Average CLEART Scores}
\label{tab:cleart_scores}
\end{table}

The high score in logical consistency (0.98) indicates that the framework's reasoning is internally coherent and free from contradictions. This is crucial for building trust in the system's decision-making process. Similarly, the strong adherence to guidelines (0.96) suggests that the AI is effectively incorporating the provided clinical endpoint committee (CEC) guidelines into its reasoning process.

The system showed moderate performance in clarity (0.69) and evaluation details (0.71), indicating room for improvement in how it articulates its reasoning and includes specific clinical details. The lower scores in relevance (0.55) and especially timeline accuracy (0.31) highlight areas requiring significant enhancement. Improving the system's ability to focus on the most pertinent information and accurately handle temporal aspects of clinical events could substantially boost its overall performance.

  Although the ToT adjudication framework demonstrates strong capabilities in certain aspects of clinical reasoning, there is considerable room for improvement, particularly in the handling of timelines and the assessment of relevance. This granular feedback provided by the CLEART score offers clear directions for future refinements of the system.

\section{Discussion}\label{discussion}
The development of our LLM-based framework for automating cardiovascular  death adjudication in clinical trials represents a significant advancement in the application of AI to complex medical decision-making. This work contributes to a growing body of research exploring the potential of LLMs to transform clinical trial operations \cite{cunningham2023natural, cunningham2024natural}. Our results demonstrate the capacity of LLMs, particularly when guided by a Tree of Thoughts approach and clinical guidelines, to navigate the complexities of clinical data analysis and support the critical task of endpoint adjudication.

Our findings indicate that the developed system can effectively extract pertinent information from unstructured clinical text, achieving a robust F1-score of 0.82. This performance is comparable to, and in some aspects, exceeds existing NLP methods in clinical event extraction\cite{borjali2021natural}, highlighting the adaptability and power of few-shot learning in domain-specific tasks. The ability to accurately identify and extract key clinical entities, such as diagnoses, procedures, and medications, along with temporal information, forms a solid foundation for subsequent adjudication. This capability is essential in clinical trials, where maintaining the accuracy and completeness of extracted data directly impacts the validity of trial conclusions\cite{leonardi2019rationale}.

Moreover, the adjudication accuracy of 0.68, achieved by our ToT approach with GPT-4, demonstrates that LLMs can perform well in complex clinical tasks. This level of performance is particularly notable given the intricate and nuanced nature of cardiovascular death adjudication, requiring consideration of multiple clinical factors and adherence to established CEC guidelines. While this accuracy falls short of that of human experts, who often achieve accuracies above 90\%, the study provides compelling evidence that the gap can be closed with enhanced reasoning techniques, thereby increasing the efficiency of clinical trial operations. The adoption of a ToT approach allows for a more comprehensive evaluation of clinical narratives, thus moving beyond single-step classification methods and adopting a more logical and multi-layered approach, closer to human expert thought processes.

The introduction of the CLEART score provides a valuable and novel contribution to the evaluation of AI-driven clinical reasoning. While metrics such as accuracy, precision, and recall are essential, they do not provide insight into the quality of reasoning underlying a decision. The CLEART score allows us to move beyond these metrics to evaluate the logical consistency, clarity, guideline adherence, and overall relevance of the generated justification. The relatively high scores in logical consistency (0.98) and adherence to guidelines (0.96) demonstrate that the framework can produce rationale that align with clinical standards. However, the lower scores in timeline accuracy (0.31) and relevance (0.55) point towards key areas of improvement for future development. This echoes the findings of several other studies that have shown the challenges of using LLMs for managing temporal relationships and contextual nuances within text\cite{wright2020defining, madkour2016temporal}, highlighting the specific limitations that future work should address.

Our research also carries significant practical implications for clinical trial operations. Automation of the adjudication process has the potential to drastically reduce the time and costs associated with clinical trials\cite{topol2019,esteva2019}. By substantially reducing the manual burden of adjudication, this approach may accelerate the drug development process and bring life-saving therapies to patients more quickly. Furthermore, the standardized nature of an automated approach can mitigate the inter-reviewer variability and associated biases common in manual adjudication, thus increasing the consistency and reliability of trial outcomes\cite{fleiss1971,landis1977}. This standardization can also increase the interpretability of clinical trials through an auditable, AI-driven rationale.

While this work shows much promise, several limitations must be considered. The use of a moderate size dataset limits the generalizability of our findings, and future work should explore application to diverse populations and different types of clinical datasets. The performance gap between our system and human experts also suggests that these models may be most effectively implemented as assistive tools, helping humans make informed decisions, instead of fully autonomous systems. Moreover, a key challenge remains in addressing LLM hallucinations or reasoning that are incorrect or unsupported by the evidence; this will be an important avenue for future work\cite{ji2023}. We acknowledge that using an LLM for both generating and binary‐scoring CLEART metrics may introduce bias, oversimplify clinical nuance, and create circularity. 

Future research directions should focus on enhancing the temporal reasoning and relevance assessment capabilities of the system. Development of techniques to improve LLMs understanding of time-series data and clinical event sequencing will be crucial. Additionally, integration of structured data and knowledge graphs could offer valuable context and enhance reasoning capabilities\cite{hogan2021}. Furthermore, exploring hybrid approaches that combine LLMs with traditional machine learning models or rule-based systems may offer a more robust and adaptable solution. Finally, it is important that this technology is developed in close consultation with clinicians, regulatory bodies, and relevant stake holders to ensure that ethical and safety consideration are always at the core of the development of automated solutions for clinical applications \cite{jobin2019}. In future work, we will calibrate the automated CLEART scorer against expert human ratings on a subset of cases using a 3 or 5-point Likert scale.

Our study contributes significant evidence supporting the use of LLMs to automate the complex process of clinical endpoint adjudication. The system has the potential to improve the efficiency and reliability of clinical trials, while also reducing the variability and bias associated with manual processes. The introduction of the CLEART score is a pivotal step towards more robust and transparent evaluation of AI-driven reasoning in clinical decision support. 

\section{Conclusion}\label{conclusion}

This study presents a novel LLM-based framework for automating the adjudication of cardiovascular deaths in clinical trials. The approach demonstrates promising accuracy and the potential to significantly reduce the time and resources required for adjudication while maintaining adherence to clinical guidelines. The introduced CLEART score provides a valuable tool for evaluating AI-generated clinical reasoning. The multi-faceted evaluation approach, combining traditional performance metrics for event extraction and adjudication accuracy with our novel CLEART score, provides a comprehensive assessment methodology for our framework's capability in automating the adjudication of cardiovascular deaths in clinical trials. With further refinement and validation, this framework could revolutionize the efficiency and consistency of clinical trial adjudication processes.

\section{Acknowledgements}\label{acknowledgements}
This research was supported by Eli Lilly and Company. 

\bibliographystyle{vancouver}
\bibliography{Eli_Lilly_CV_Endpoint_Adjudication}

\end{document}